\documentclass[10pt, conference]{IEEEtran}
\IEEEoverridecommandlockouts
\usepackage{cite}
\usepackage{amsmath,amssymb,amsfonts}
\usepackage{algorithmic}
\usepackage{graphicx}
\usepackage{textcomp}
\usepackage{amssymb}
\usepackage{pifont}
\usepackage{booktabs}
\usepackage{graphicx}

\usepackage[utf8]{inputenc} 
\usepackage[T1]{fontenc}    
\usepackage{hyperref}       
\usepackage{url}            

\usepackage{amsfonts}       
\usepackage{nicefrac}       
\usepackage{microtype}      

\usepackage{multirow}
\usepackage{multicol}
\usepackage{pifont}
\usepackage{comment}
\usepackage{adjustbox}
\usepackage{color}
\usepackage{enumitem}
\usepackage{amsmath}
\usepackage{caption}
\usepackage{diagbox}
\usepackage{enumitem}
\usepackage{booktabs}
\usepackage{multirow}
\usepackage{multicol}
\usepackage{xcolor}
\def\BibTeX{{\rm B\kern-.05em{\sc i\kern-.025em b}\kern-.08em
    T\kern-.1667em\lower.7ex\hbox{E}\kern-.125emX}}
\begin{document}

\title{Spatial 3D-LLM: Exploring Spatial Awareness in 3D Vision-Language Models}
\def\confName{Spatial 3D-LLM}

\author{
\thanks{*Corresponding author, zhangchao@bdnrc.org.cn}
\IEEEauthorblockN{
        Xiaoyan Wang$^{1}$,
		Zeju Li$^{1,2\dagger}$,\thanks{$\dagger$ This work was done during the internship at BDNRC.} 
        Yifan Xu$^{1}$,
        Jiaxing Qi$^{1}$,
        Zhifei Yang$^{1}$, 
        Ruifei Ma$^{1}$,
        Xiangde Liu$^{1}$,
		Chao Zhang$^{1 *}$	
}
	\IEEEauthorblockA{$^1$\textit{Beijing Digital Native Digital City Research Center}, $^2$\textit{The Chinese University of Hong Kong}}

}

\maketitle

\begin{abstract}
New era has unlocked exciting possibilities for extending Large Language Models (LLMs) to tackle 3D vision-language tasks. However, most existing 3D multimodal LLMs (MLLMs) rely on compressing holistic 3D scene information or segmenting independent objects to perform these tasks, which limits their spatial awareness due to insufficient representation of the richness inherent in 3D scenes. To overcome these limitations, we propose Spatial 3D-LLM, a 3D MLLM specifically designed to enhance spatial awareness for 3D vision-language tasks by enriching the spatial embeddings of 3D scenes. Spatial 3D-LLM integrates an LLM backbone with a progressive spatial awareness scheme that progressively captures spatial information as the perception field expands, generating location-enriched 3D scene embeddings to serve as visual prompts. Furthermore, we introduce two novel tasks: 3D object distance measurement and 3D layout editing, and construct a 3D instruction dataset, \textbf{MODEL}, to evaluate the model's spatial awareness capabilities. Experimental results demonstrate that Spatial 3D-LLM achieves state-of-the-art performance across a wide range of 3D vision-language tasks, revealing the improvements stemmed from our progressive spatial awareness scheme of mining more profound spatial information. Our code is available at \url{https://github.com/bjshuyuan/Spatial-3D-LLM}.
\end{abstract}

\begin{IEEEkeywords}
3D-LLM, Spatial perception and reasoning, Progressive spatial awareness scheme, Dataset
\end{IEEEkeywords}

\section{Introduction}
\label{sec:intro}
In recent years, Vision-Language Models (VLMs)\cite{li2022blip,llava} have rapidly advanced, with 2D Multimodal Large Language Models (MLLMs) demonstrating remarkable capabilities in understanding complex visual scenes. 
Concurrently, much success of developing 3D MLLMs has been achieved on 3D scene understanding\cite{3dllm,3dmit,xu2025argus,huang2023chat3dv2,huang2024leo}. 
3D spatial awareness encompasses the perception of spatial states, such as locations and distances, as well as spatial reasoning and generation derived from this perception, including embodied planning and spatial layout editing.
While diving into 3D world, 3D spatial awareness is one of the keys for 3D MLLMs to perform downstream tasks, such as robotics\cite{gao2024physically}, virtual reality\cite{konenkov2024vr} and interior design\cite{yang2024llplace}.

To enable VLMs to perceive and comprehend the 3D world, most existing 3D MLLM architectures incorporate a 3D vision encoder to extract 3D features and align them with an LLM\cite{3dllm, huang2023chat3dv2,3dmit}.
Current methods\cite{huang2023chat3dv2,huang2024leo} primarily focus on segmented object attributes, overlooking strategies for precise 3D location perception.
Approaches like \cite{3dllm} and \cite{chen2024ll3da} utilize the Q-former\cite{li2022blip} module to extract instruction-related information from 3D scene embeddings, forming the input for 3D MLLMs. 
However, the extracted embeddings are overly aligned with the instructions, failing to fully capture the spatial concepts of 3D scenes.
Existing works \cite{huang2023chat3dv2, huang2024leo, chen2024ll3da,chen2024grounded3dllm} still lack effective perception of 3D spatial relations and precise location generation, which are fundamental for spatial reasoning and generation.
In 3D scenes, spatial information exists naturally at various levels, including that of individual objects, object groupings, and entire architectures.
Consequently, most of existing 3D MLLMs either compress holistic 3D scene information or segment independent objects for 3D vision-language(3D VL) tasks, limiting their spatial awareness due to insufficient representation of the richness inherent in 3D scenes. 

Given the aforementioned concerns regarding the inadequate exploitation of spatial information in existing 3D MLLMs, we propose \textbf{Spatial 3D-LLM}, a 3D MLLM aimed at improving capabilities of spatial awareness for 3D VL tasks by enriching the spatial embeddings of 3D scenes.
Spatial 3D-LLM integrates a frozen 3D scene encoder, an LLM backbone, and a meticulously designed progressive spatial awareness scheme that includes intra-referent clustering and abstraction, inter-referent message passing, and contextual referent-scene interactions.
This spatial awareness visual referent evolution begins with relation-based clustering. It then continues with inter-referent message passing to model spatial distribution based on the distances between different referents.
Finally, it encompasses broader contextual information by considering the interactions between referents and the surrounding environment.
This stepwise scheme progressively captures spatial information as the perception field expands, injecting location-enriched spatial knowledge into the 3D scene embeddings.
These enhanced embeddings serve as visual prompt for end-to-end instruction tuning, eliminating the task-specific optimizations.

Considering spatial awareness from the perspective of tasks and datasets, several works\cite{cheng2024spatialrgpt,chen2024spatialvlm} have improved image-based spatial reasoning capabilities by generating large-scale spatially-aware training data. 
They hypothesize that VLMs’ limited spatial reasoning capability is due to the lack of 3D spatial knowledge in training data\cite{cheng2024spatialrgpt, chen2024spatialvlm}.
Those generated question answering datasets are mainly related to estimating object pair relationships and metric measurements. 
Existing 3D instruction following datasets\cite{li2023m3dbench,lyu2024mmscan} 
support a variety of spatial tasks, including visual question answering, visual grounding, and spatial relationships estimation. 
However, these datasets mainly concentrate on perceiving coarse-grained and abstract object relationships while leaving fine-grained measurement unexplored. 
Moreover, they typically focus on local object interactions, neglecting the utilization of commonsense knowledge of object-scene spatial information.

In light of the mentioned deficiencies in existing 3D instruction datasets, we propose two novel tasks: 3D object distance measurement and 3D layout editing in 3D scenes, to evaluate the spatial awareness capabilities of 3D MLLMs. 
We construct a 3D instruction dataset called \textbf{M}easure \textbf{O}bject \textbf{D}istance and \textbf{L}ayout \textbf{E}diting (MODLE) that is furnished with 263K vision-language annotations specifically targeted towards these tasks.
Inferring precise distances between objects enhances fine-grained spatial perception, while performing object placement and movement in a 3D scene fosters a deeper understanding of object-scene spatial information, accumulating commonsense knowledge for downstream tasks. 

In summary, our contributions are as follows:
\begin{itemize}[leftmargin=*]
    \item We propose two novel location-related tasks in 3D scenes, namely 3D object distance measurement and 3D layout editing.
    We construct \textbf{MODLE}, a 3D instruction dataset furnished with 263K vision-language annotations towards these tasks.
    \item We present \textbf{Spatial 3D-LLM}, a 3D MLLM that improves 3D spatial awareness capabilities by enriching the spatial embeddings of 3D scenes.
    Spatial 3D-LLM features a progressive spatial awareness scheme that captures spatial information as the perception field expands, injecting location-enriched spatial knowledge into the 3D scene embeddings. 
    \item Experimental results demonstrate that our method achieves state-of-the-art performance across diverse 3D VL tasks, especially those concerning locations and spatial relationships.
\end{itemize}

\section{Related Work}
\label{sec:related}

\subsection{Spatial Awareness in 3D Vision-Language Tasks}
Diverse 3D VL tasks pose disparate demands on a model's capability of spatial perception and reasoning within 3D environments.
For instance, \textbf{3D Visual Question Answering (3D-VQA)} \cite{azuma2022scanqa,ma2022sqa3d} primarily rely on understanding the holistic scene to provide answers or descriptions, without delving deeply into object-to-object spatial configurations.
\textbf{3D Visual Grounding(3D-VG)} \cite{zhang2023multi3drefer,chen2020scanrefer} focus on identifying and locating specific objects within the 3D space.
Additionally, \textbf{3D Dense Captioning} \cite{chen2021scan2cap} involves generating detailed descriptions for various regions or objects in a 3D scene, requiring a strong grasp of how objects are positioned and interact within their spatial context.

Existing 3D VL tasks primarily focus on perceiving coarse-grained and abstract object relationships, coupled with concentrating on local object interactions.
Our newly proposed tasks enhance fine-grained spatial perception and accumulate commonsense knowledge for downstream tasks, thereby advancing the capability of spatial awareness.

\subsection{Spatial Learning in 3D Multimodal LLMs}
Recent advancements in 3D MLLMs\cite{3dllm,chen2024ll3da} have explored a variety of spatial learning paradigms.
These architectures typically comprise 3D vision perceptrons, projectors, and LLM backbones.
3DLLM\cite{3dllm} introduced location special tokens to better capture 3D spatial information, enabling models to output 3D coordinates.
LL3DA\cite{chen2024ll3da} used clicks and boxes as visual prompts to interact with 3D embeddings and generate spatial queries.
SpatialRGPT\cite{cheng2024spatialrgpt} enhanced region-level spatial reasoning in VLMs by improving regional information representation and spatial knowledge acquisition.
Chat-3D v2\cite{huang2023chat3dv2} segmented scenes into objects, mapped each with an index, and used special tokens to capture 3D attributes and spatial relations.
Grounded 3DLLM\cite{chen2024grounded3dllm} introduced special noun phrase tokens to reference 3D scenes and let models process 3D-textual data sequences.
Most existing 3D MLLMs rely on holistic 3D scene information or specifically designated regions, missing multi-level location-based information.

Distinguished from current approaches, our method explores a progressive spatial awareness scheme that incorporates intra-referent clustering and abstraction, inter-referent message passing, and contextual referent-scene interactions, injecting richer spatial knowledge into the 3D scene embeddings.

\section{Datasets}
We propose 3D object distance measurement and 3D layout editing tasks for improving 3D spatial perception capabilities of our \textbf{Sptial 3D-LLM}, and accumulating commonsense knowledge for downstream tasks. 
Hence, we construct a visual language instruction dataset for these two tasks, namely \textbf{MODLE}.
Statistics for the datasets are provided in Table \ref{tab:dataset_description}, with relevant evaluation metrics and examples are shown in Support Material.

\begin{table}[htbp]
\caption{\textbf{Statistic results of our proposed MODLE dataset.} [BOX] represents the 3D bounding box of an object, and [DIS] represents the distance value between objects.}
\vspace{-8pt}
\begin{center}
\resizebox{\linewidth}{!}{
\begin{tabular}{lcccccc}
\toprule
\multirow{2}{*}{Tasks} & \multirow{2}{*}{\#3D Scan} & \multicolumn{2}{c}{\#Language} & \multirow{2}{*}{Object-level} & \multirow{2}{*}{Text Instructions} & \multirow{2}{*}{Output Type} \\ 
\cmidrule(lr){3-4}
                       &                           & Train & Val                        &                                  &                                  &                              \\
\midrule
3D Object Distance Measurement & 0.7K & 171K  & 2K  & Multi  & <obj caption> & [BOX], [DIS] \\
\midrule
3D Object Movement           & 0.7K & 36K   & 9K  & Single & <obj caption> & [BOX] \\
3D Object Placement          & 0.69K & 34K   & 9K  & Single & <obj caption> & [BOX] \\
\bottomrule
\end{tabular}
}
\label{tab:dataset_description}
\end{center}
\end{table}

\vspace{-3pt}
\subsection{3D Object Distance Measurement Task}

\label{3D_Distance_Measurement}
This task focuses on inferring 3D spatial distance between two objects within 3D scenes.
We create 173K text-location pairs.
Questions are made with manually defined templates, generating synthetic data by filling in object descriptions sourced from ScanRefer\cite{chen2020scanrefer} dataset.
Answers are derived from the actual 3D bounding box coordinates of the objects.
We introduce Interaction Tokens to distinguish between coordinate information and distance values in the output.
Coordinates are put within \texttt{<loc></loc>} tokens, and distances are marked with \texttt{<gap></gap>} tokens.

\subsection{3D Layout Editing Task}
\label{Scene_Edit}
This task demands the model have 3D layout editing capabilities. We design two subtasks: object movement and placement. Unlike the 3D-VG task that grounds an object in the scene, 3D layout editing requires a precise understanding of 3D spatial positions for predicting new object positions.

For the object movement task, the model is required to relocate an object in the scene based on its description and an editing instruction.
We define a template including the object description and movement instructions to construct the dataset with 45K text-location pairs.
The object descriptions come from the ScanRefer dataset, and the movement instructions are randomly generated.
In the object placement task, the model needs to understand the holistic scene and accurately place an object of a specified size within the scene layout.
We created 33K sub-scenes from the ScanNet\cite{dai2017scannet} dataset, each with 3 to 8 objects.
During training and evaluation, we randomly mask one object from each sub-scene and require the model to predict a reasonable spatial position.

\section{Method}
We propose Spatial 3D-LLM, a 3D MLLM designed for comprehensive spatial awareness, including spatial perception, reasoning, and generation. 
The main pipeline of Spatial 3D-LLM is illustrated in Figure \ref{method}.
Spatial 3D-LLM incorporates a frozen 3D scene encoder, an LLM backbone, and a progressive spatial awareness scheme. 
In the following sections, we will provide a detailed explanation of each component.

\begin{figure*}[t]
\centering
\includegraphics[width=0.95\textwidth]{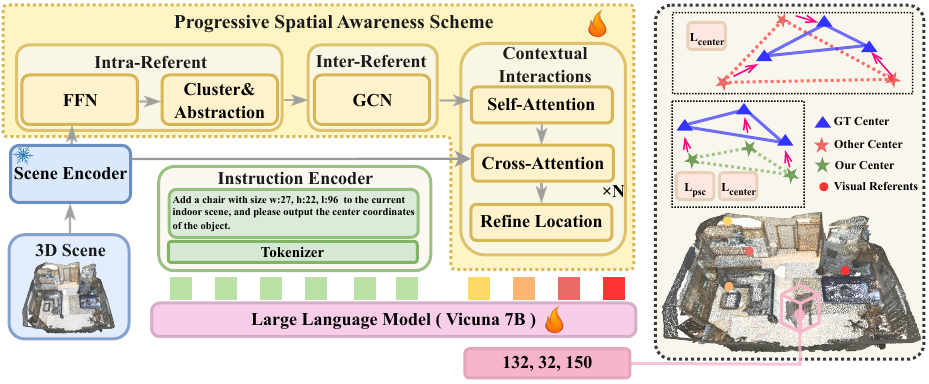} 
\label{method}
\caption{\textbf{The model architecture of Spatial 3D-LLM.} It includes a frozen 3D scene encoder, an LLM backbone, and a meticulously designed progressive spatial awareness scheme that incorporates intra-referent clustering and abstraction, inter-referent message passing, and contextual referent-scene interactions, generating location-enriched 3D scene embeddings.}
\label{method}
\end{figure*}

\subsection{Scene Encoder}
To handle the point clouds in the 3D scene, we utilize PointNet++ \cite{qi2017pointnet++} as our scene encoder, which employs a hierarchical neural network to convert the unordered point set into an unordered set of point features. To represent the input 3D scene, the scene encoder outputs 1,024 point tokens, $F_{\text{enc}}=[p_{\text{enc}}, f_{\text{enc}}] \in \mathbb{R}^{1,024 \times (3+256)}$, containing scene features $f_{\text{enc}}$ for 256 dimensions and coordinates $p_{\text{enc}}$ for 3 dimensions.

\subsection{Progressive Spatial Awareness Scheme}
\label{sptaial_tran}
The architecture of the Progressive Spatial Awareness Scheme comprises intra-referent module, inter-referent module, and contextual interactions module. Next, we will explain the details of each part.
 
\subsubsection{Intra-Referent}
To capture point-to-point relational information within the scene, we propose the Intra-referent module, which comprises Feedforward Neural Network (FFN) layers and a Cluster\&Abstraction layer. 
Specifically, we sample 256 points from the encoded set of 1024 scene points using Farthest Point Sampling (FPS), resulting in seed points $F_{\text{seed}}=[p_{\text{seed}}, f_{\text{seed}}] \in \mathbb{R}^{256 \times (3+256)}$, following VoteNet\cite{votenet}. Next, 3D spatial offset \(\Delta \mathbf{p}_{vote}\) is predicted from the seed point feature \({f}_{seed}\) by means of FFN layers. With the aim of adjusting seed point location to align with the centers of object, as indicated by:
\vspace{-4pt}
\[
{p}_{vr} = {p}_{seed} + FFN({f}_{seed}) = {p}_{seed} + \Delta \mathbf{p}_{vote}
\]
\vspace{-1pt}
we aggregate local information from $F_{\text{seed}}$ for \({p}_{vr}\) with a Cluster \& Abstraction layer, to form the visual referent reprention, as $F_{\text{vr}}=[p_{\text{vr}}, f_{\text{vr}}]$. For each visual referent location \({p}_{vr}\), its neighboring points are grouped to form local regions, and the features of these regions from  \({f}_{enc}\) are abstracted by pooling, mapping the set of points to a feature vector, as visual referent embedding \({f}_{vr}\):
\vspace{-4pt}
\[
{f}_{vr} = Pooling( Cluster[{p}_{vr}, {f}_{enc}])
\]
\vspace{-1pt}
This process generates visual referent representation $F_{\text{vr}}=[p_{\text{vr}}, f_{\text{vr}}] \in \mathbb{R}^{256 \times (3+256)}$, which encapsulates the internal point-to-point relationships within the local region.

\subsubsection{Inter-Referent}
We contend that relying solely on the feature information from local region is insufficient for adequately representing 3D scenes. Consequently, we propose the Inter-Referent module, which employs a Graph Convolutional Network (GCN) model for message propagation to facilitate the modeling of global spatial distribution among visual referents, particularly focusing on the implicit relationships between these referents within the scene.

In this approach, we model the graph nodes using \(f_{\text{vr}}\), with edges defined based on the distances between visual referents locations \(p_{\text{vr}}\). The forward propagation for each layer of the GCN can be expressed as follows: ${H}^{(l+1)} = \sigma \left( {A} {H}^{(l)}{W}^{(l)} \right)$, where ${H}^{(l)}$ represents the node embeddings at layer $l$, ${A}$ is the adjacency matrix capturing the spatial connections between nodes, ${W}^{(l)}$ is the weight matrix of the GCN layer, and $\sigma$ is the activation function.

By iteratively training the GCN, we learn an enhanced representation for each visual referent that captures both its local features and the global spatial context from its neighboring visual referents. The output of the layers of GCN is a refined visual referent representation $F_{\text{vr-gcn}}=[p_{\text{vr}}, f_{\text{vr}}] \in \mathbb{R}^{256 \times (3+256)}$, which is obtained through inter-referent message passing. 

\subsubsection{Contextual Interactions}
To achieve contextual interactions among visual referents and the global scene, we introduce the Contextual Interactions module, which employs multiple blocks of self-attention, cross-attention, and refine-location layer to obtain spatially aware representations of referents. In detail, visual referent $F_{\text{vr-gcn}}$ processed by Inter-referent module, along with the scene features $f_{\text{enc}}$ undergo self-attention and cross-attention layers, resulting in the updated features $F_{\text{vr-enc}}$, which can be claimed as scene-aware visual referent representations. They effectively incorporating object-specific features and spatial positional information, thereby enhancing its comprehensiveness and achieving spatial scene awareness.

\paragraph{Refine-Location.}
To further improve the precision of visual referent location predictions, we introduce the \textit{Refine-Location} layer, designed to refine the spatial positioning of referents by minimizing the relative distance to their ground truth coordinates. 

This module comprises multiple layers of FFN that learn positional offsets to adjust the locations of visual referents, aligning them more closely with the object's coordinate center. We define a visual referent's ground truth location as the centroid of the nearest object. Consequently, supervision derives from these ground truth locations, aiming to minimize both center distance and pairwise distance between predicted and actual visual referent pairs, quantified through \textit{center loss} ($\mathcal{L}_{\text{center}}$) and the \textit{pairwise spatial constraint loss} ($\mathcal{L}_{\text{psc}}$) which are computed as:
\vspace{-4pt}
\[
    \mathcal{L}_{\text{center}} = \frac{1}{M} \sum_{i=1}^{M} \| q_{\text{vr}}^{(i)} - q_{\text{gt}}^{(i)} \|_2, 
    \mathcal{L}_{\text{psc}} = \frac{1}{N} \sum_{i=1, j=1}^{N} \| k_{\text{vr}}^{(ij)} - k_{\text{gt}}^{(ij)} \|_2,    
\]
\vspace{-1pt}
where \( M \) is the number of visual referent, \( N \) is the number of visual referent pairs, $q_{\text{vr}}^{(i)}$ and $q_{\text{gt}}^{(i)}$ are denoted as the coordinates of the predicted and ground truth visual referent, \(k_{\text{vr}}^{(ij)} \) is the predicted distance of visual referent pairs \( (i, j) \), \( k_{\text{gt}}^{(ij)} \) is the corresponding ground truth distance. This loss penalizes the Euclidean distance between the predicted and ground truth distance, encouraging the model to predict more accurate visual referent locations.

By using a progressive visual referent evolution approach that enhances the perception field with spatial information, learned 3D scene embeddings effectively capture location-enriched spatial knowledge. This allows our model to excel in spatial position perception and improve its ability for spatial awareness in 3D vision-language tasks.

\subsection{Spatial 3D-LLM training}
For representing 3D point coordinates occuring the text, following LL3DA\cite{chen2024ll3da}, we normalize the point cloud coordinates into discrete unsigned integers within the range [0-255]. 
This representation is distinguished by special token \texttt{<loc></loc>}, which help differentiate the spatial coordinates from other data.

To integrate the visual prompt, denoted as ${F}_{visual}$, which contains both visual referent features ${f}_{vr}$ and location representation ${p}_{vr}$ into the large language model (LLM). We introduce a trainable projector consist multi-layers of FFN to align ${F}_{visual}$ = $[{f}_{vr},{p}_{vr}]$ as \textbf{Visual Prompt} within the language space of LLM, allowing the model to process 3D spatial information alongside natural language input.

We use the instruction tuning paradigm for training our Spatial 3D-LLM. With VL understanding and VL grounding tasks consist of the training dataset, we get the
loss, denoted as $\mathcal{L}_{\text{LLM}}$, is computed based on the model’s performance on these tasks.
In addition to $\mathcal{L}_{\text{LLM}}$, we also introduce $\mathcal{L}_{\text{psc}}, \mathcal{L}_{\text{center}}$ to get more precise coordinates generation and understanding. Thus, the overall optimization objective is the sum of these three losses:
\vspace{-4pt}
\[
    \mathcal{L}_{\text{total}} =  \mathcal{L}_{\text{LLM}} + \alpha_1 \mathcal{L}_{\text{psc}} + \alpha_2 \mathcal{L}_{\text{center}},
\]
\vspace{-4pt}
where $\mathcal{L}_{\text{LLM}}$ is the loss associated with the instruction tuning tasks for our based LLM, $\alpha_1$ and $\alpha_2$ are weighting factors.
By optimizing this combined loss, Spatial 3D-LLM learns both precise spatial information through the spatial loss and instruction-following capabilities via our based LLM instruction tuning loss.
\vspace{-4pt}
\section{Experiments}
\vspace{-2pt}
\subsection{Datasets and Implementation Details}
\vspace{-2pt}
\textbf{Datasets.} 
To evaluate the performance of our model, we require 3D scene point clouds along with a visual-language task dataset. For the 3D scene input, we utilize ScanNet\cite{dai2017scannet}, a real 3D indoor scene dataset that includes 1,201 training scenes and 312 testing scenes. For visual-language data, we incorporate Scan2Cap\cite{chen2021scan2cap}, ScanQA\cite{azuma2022scanqa}, SQA3D\cite{ma2022sqa3d}, and embodiedQA\cite{3dllm} from 3D-LLM for training and evaluation of visual-language understanding tasks. Additionally, we use ScanRefer\cite{chen2020scanrefer} and Multi3DRefer\cite{zhang2023multi3drefer} for single- and multi-object grounding, and leverage proposed distance measurement, object movement, and object placement tasks for precise spatial position perception and generation.

\textbf{Implementation Details.}
We initialize the weights of the 3D scene encoder using the pre-trained Vote2Cap-DETR\cite{Vote2cap-detr++}. The large language model utilizes the pre-trained Vicuna-7B and implement LoRA for instruction-tuning. During the training process, we jointly train the progressive spatial awareness scheme and the LoRA parameters across all task datasets. We employ AdamW as the optimizer, with a learning rate between $10^{-4}$ and $10^{-7}$ and a weight decay of 0.1. All experiments are conducted on eight A100 GPUs within one day.

\subsection{Comparison with SOTA models}
To evaluate the spatial awareness capabilities of our model, we present the evaluation results on two types of tasks: 3D vision-language understanding tasks and 3D vision-language grounding tasks. The qualitative results are shown in Support Material. It is important to highlight that all tasks were trained concurrently throughout the training process, and for each evaluation task, the evaluation metrics are derived from the same model weights.

\subsubsection{3D Vision-Language Understanding}
As shown in Table \ref{tab: understanding-compare} reporting the explicit performance on the Scan2Cap, ScanQA and SQA3D tasks. We categorize the existing methods into three groups: task-specific models tailored for downstream tasks; task-specific fine-tuned approaches that involve pretraining a unified 3D backbone followed by subsequent fine-tuning for specific tasks; and generalist models capable of comprehending a range of 3D vision-language tasks.

\paragraph{Analysis}
Table \ref{tab: understanding-compare} shows that our method surpasses most methods in terms of CIDEr\cite{vedantam2015cider}, BLEU-4\cite{papineni2002bleu}, METEOR\cite{banerjee2005meteor}, and ROUGE\cite{lin2004rouge} across all three tasks. 
In the Scan2Cap task, which involves localizing and generating descriptive captions for objects in 3D scenes, our method achieves a significantly higher CIDEr score, reflecting its ability to generate more accurate and contextually relevant captions. Similarly, in the ScanQA task, which tests the model’s ability to answer questions with more semantic diversity about 3D scenes, our method shows notable improvements across all metrics, particularly in CIDEr and BLEU-4.
Furthermore, in the SQA3D task, which involves answering situated questions in complex 3D environments, Spatial 3D-LLM once again excels, showcasing its robustness in understanding both spatial and linguistic nuances. Overall, our method consistently surpasses other models in key performance metrics, demonstrating its advanced spatial reasoning capabilities and comprehensive understanding of 3D scenes.

\begin{table}[!tb]
\caption{\textbf{Quantitative comparison with SOTA models on 3D VL understanding tasks.} ``C'' stands for ``CIDEr'', ``B-4'' for ``BLEU-4'', ``M'' for ``METEOR'', ``R'' for ``ROUGE'', and ``EM@1'' for top-1 exact match. The n-gram metrics for Scan2Cap are governed by IoU@0.5. $^\dagger$ indicates answering questions via prompting GPT-3 with the generated scene caption.}
\label{tab: understanding-compare}
\resizebox{\linewidth}{!}{
\begin{tabular}{lccccccccc}
    \toprule
     & \multicolumn{4}{c}{Scan2Cap} & \multicolumn{4}{c}{ScanQA} & SQA3D \\
     \cmidrule(lr){2-5} \cmidrule(lr){6-9} \cmidrule(lr){10-10} 
     & C & B-4 & M & R & C & B-4 & M & R & EM@1 \\
    \midrule
    \multicolumn{1}{l}{\small\textit{\textbf{Task-specific models}}} \\
    Scan2Cap\cite{chen2021scan2cap}        & 35.2 & 22.4 & 21.4 & 43.5 & - & - & - & - & \hspace{2pt} 41.0$^\dagger$ \\
    Vote2Cap-DETR\cite{Vote2cap-detr++}   & 67.6 & \textbf{37.1} & 26.9 & \textbf{55.6} & - & - & - & - & - \\
    ScanRefer+MCAN\cite{chen2020scanrefer}  & - & - & - & - & 55.4 & 7.9 & 11.5 & 30.0 & - \\
    ScanQA\cite{azuma2022scanqa}          & - & - & - & - & 64.9 & 10.1 & 13.1 & 33.3 & 47.2 \\
    \midrule
    \multicolumn{1}{l}{\small\textit{\textbf{Task-specific fine-tuned}}} \\
    3D-VisTA\cite{3dvista}        & 66.9 & 34.0 & \textbf{27.1} & 54.3 & 69.6 & 10.4 & 13.9 & 35.7 & \textbf{48.5} \\
    3D-LLM (FlanT5)\cite{3dllm} & - & - & - & - & 69.4 & 12.0 & 14.5 & 35.7 \\
    Chat-3D v2\cite{huang2023chat3dv2}      & - & - & - & - & 77.1 & 7.3 & 16.1 & \textbf{40.1} & -  \\ 
    LL3DA\cite{chen2024ll3da}           & 65.2 & 36.8 & 26.0 & 55.1 & 76.8 & 13.5 & 15.9 & 37.3 \\ 
    \midrule
    \multicolumn{1}{l}{\small\textit{\textbf{Generalist models}}} \\
    LL3DA\cite{chen2024ll3da}           & 63.0 & 36.0 & 25.7 & 54.7 & 75.7 & 13.3 & 15.4 & 37.0 & - \\
    Grounded 3D-LLM\cite{chen2024grounded3dllm} & 70.6  & 35.5 & - & - &  72.7 & 13.4 &  -  &  -  &  -  \\
    Spatial 3D-LLM (Ours)  & \textbf{72.2} & 34.6 & 23.1 & 54.3 & \textbf{82.5} & \textbf{13.9} & \textbf{16.8} & 39.1 & 46.2 \\
    \bottomrule
\end{tabular}
}
\end{table}

\begin{table}[!tb]
\centering
\caption{\textbf{Quantitative comparison with SOTA models on 3D VL grounding tasks.} [BOX] indicates models that output 3D bounding boxes, while [ID] indicates models that output individual object IDs. ReGround3D 3D-LLM refers to the reproduced 3D-LLM results from the ReGround3D model.}
\label{tab: grounded-compare}
\resizebox{\linewidth}{!}{
\begin{tabular}{lccccc}
\toprule
\multirow{2}{*}{} & \multirow{2}{*}{Output Type} & \multicolumn{2}{c}{ScanRefer Grd.} & \multicolumn{2}{c}{Multi3DRef Grd.}  \\ 
\cmidrule{3-4} \cmidrule(lr){5-6}
                       &          & Acc@0.25 & Acc@0.5 & F1@0.25 & F1@0.5             \\ 
\midrule
ScanRefer\cite{chen2020scanrefer}              & [BOX]    & 37.3     & 24.3    & -       & -                  \\
M3DRef-CLIP\cite{zhang2023multi3drefer}            & [BOX]    & 51.9     & \textbf{44.7}    & 42.8    & 38.4              \\
LLM-Grounder\cite{Yang2023LLMGrounderO3}           & [BOX]    & 17.1     & 5.3     & -       & -                  \\
Chat-3D v2\cite{huang2023chat3dv2}             & [ID]     & 35.9     & 30.4    & -       & -               \\
ReGround3D 3D-LLM\cite{Zhu2024ScanReasonE3}      & [BOX]    & 33.1     & 28.7    & -       & -              \\
Grounded 3D-LLM\cite{chen2024grounded3dllm}        & [ID]     & \textbf{47.9}     & 44.1    & 45.2    & 40.6               \\
\midrule
Spatial 3D-LLM (Ours)          & [BOX]    & 44.3     & 37.2    & \textbf{48.3}    & \textbf{41.2}               \\
\bottomrule
\end{tabular}
}
\end{table}

\subsubsection{3D Vision-Language Grounding}
Table \ref{tab: grounded-compare} presents a quantitative comparison between our method and several SOTA models on 3D vision-language grounding tasks, evaluated on the ScanRefer\cite{chen2020scanrefer} and Multi3DRef\cite{zhang2023multi3drefer} benchmarks. We report the evaluation metrics of Acc@0.25 and Acc@0.5 for visual grounding on ScanRefer, and F1@0.25 and F1@0.5 for multi-object visual grounding on Multi3DRef.

\paragraph{Analysis}
Table \ref{tab: grounded-compare} shows that our method demonstrates competitive performance across both tasks. Specifically, in the ScanRefer visual grounding task, our approach achieves the Acc@0.25 score of 44.3\% and Acc@0.5 of 37.2\%, closely matching the performance of Grounded 3D-LLM and outperforming other several baselines. In the Multi3DRef visual grounding task, our model achieves the F1@0.5 score of 41.2\% and F1@0.25 score of 48.3\%, which outperforms than other baselines such as ReGround3D and Grounded 3D-LLM.

These results demonstrate the effectiveness of our method, particularly in multi-object grounding scenarios. While our model slightly lags behind the top-performing models in terms of overall accuracy in the ScanRefer task, it excels in the Multi3DRef task, showing its strength in handling complex spatial relationships across multiple objects. The consistent performance across different metrics highlights the robustness and versatility of our approach in 3D Vision-Language grounding tasks. Notably, our model directly outputs precise 3D bounding boxes for object localization, offering a significant advantage over similar previous SOTA methods like ReGround3D 3D-LLM.

\vspace{-2pt}
\subsection{Ablation Studies}
To further evaluate the effectiveness of the implementation of progressive spatial awareness scheme in enhancing the performance of our Spatial 3D-LLM, we evaluate our model on both existing tasks and proposed benchmark to conduct ablation studies. 

\textbf{Analysis of ablation studies of different components.}
Table \ref{tab: ablation-components} presents the results of ablation studies on key components: Intra-Referent(\textbf{C1}), Inter-Referent(\textbf{C2}) and Contextual Interactions(\textbf{C3}). Our model (C1 + C2) outperforms C1 alone, demonstrating the effectiveness of the Inter-Referent Module(\textbf{C2}). This is attributed to its ability to learn the implicit relationships between visual referents.
Our full model (C1 + C2 + C3) consistently outperforms both the C1 model and the (C1 + C2) model, demonstrating the effectiveness of the Contextual Interaction module (\textbf{C3}) in learning referent-scene interactions.
In Scan2Cap, it achieves the highest CIDEr and BLEU-4 scores. Similarly, for Multi3DRef, our model outperforms the C1 and C2 variants.
The full model also shows superior performance in object editing tasks, achieving higher accuracy in both movement and placement tasks. For the measurement task, our model demonstrates lower mean absolute relative error (X/Y/Z-mARE) compared to the alternatives, emphasizing the contribution of each component to spatial understanding. 
Furthermore, our proposed progressive spatial awareness scheme has a significant impact on performance, providing superior feature extraction that supports complex spatial reasoning.

\begin{table}
\centering
\caption{\textbf{Ablation studies of different components.}\textbf{C1} represents Intra-Referent module, \textbf{C2} represents Inter-Referent module, and \textbf{C3} represents Contextual Interactions module.}
\label{tab: ablation-components}
\resizebox{\linewidth}{!}{
\begin{tabular}{ccccclcclclclc} 
\toprule
\multirow{2}{*}{C1} & \multirow{2}{*}{C2} & \multirow{2}{*}{C3} & \multicolumn{2}{c}{Scan2Cap} &  & \multicolumn{2}{c}{Multi3DRef Grd} &  & Movement &  & Placement &  & Measurement     \\ 
\cmidrule{4-5}\cmidrule{7-8}\cmidrule{10-10}\cmidrule{12-12}\cmidrule{14-14}
&                     &                     & C     & B-4                &  & F1@0.25 & F1@0.5                   &  & Acc@0.5   &  & Acc@0.5  &  & X/Y/Z-mARE@0.5           \\ 
\midrule
\ding{51}             &                     &                     & 52.1 & 32.5              &  & 30.4    & 15.9                  &  & 31.7     &  & 46.8    &  & 7.5/8.7/7.2  \\
\ding{51}      &        &         \ding{51}    & 67.1 & 33.3              &  & 43.4    & 37.0                    &  & 34.2      &  & 59.7    &  & 2.6/2.1/4.5  \\
 \ding{51}                &          \ding{51}            &         \ding{51}           &  \textbf{72.2}    & \textbf{34.6}        &  & \textbf{48.3}   & \textbf{41.2}       &  & \textbf{40.3}    &  & \textbf{66.4}    &  & \textbf{2.0/1.4/2.4}   \\
\bottomrule
\end{tabular}}
\end{table}

\vspace{-4pt}
\section{Conclusion}
In this paper, we have presented Spatial 3D-LLM, a multi-modal LLM for 3D scene understanding and spatial perception, which could fully exploit the spatial awareness within 3D scenes. By carefully designing a progressive spatial awareness scheme within our framework, Spatial 3D-LLM captures location-enriched spatial knowledge in the 3D scene embeddings. Moreover, two novel tasks including 3D object distance measurement and 3D layout editing are proposed to evaluate fine-grained spatial awareness capabilities of our framework. The experimental results verify our Spatial 3D-LLM's capability in 3D scene understanding and spatial perception. Our future work includes expanding the diversity of the training datasets to encompass more complex and varied 3D scenes, which would enhance the model's generalizability. Additionally, we aim to investigate methods to improve real-time performance without compromising accuracy for practical applications such as robotics and augmented reality.

\bibliographystyle{IEEEbib}
\bibliography{icme2025}

\begin{thebibliography}{10}

\bibitem{li2022blip}
Junnan Li, Dongxu Li, Caiming Xiong, and Steven Hoi,
\newblock ``Blip: Bootstrapping language-image pre-training for unified vision-language understanding and generation,''
\newblock in {\em International conference on machine learning}. PMLR, 2022, pp. 12888--12900.

\bibitem{llava}
Haotian Liu, Chunyuan Li, Qingyang Wu, and Yong~Jae Lee,
\newblock ``Visual instruction tuning,''
\newblock {\em Advances in neural information processing systems}, vol. 36, 2024.

\bibitem{3dllm}
Yining Hong, Haoyu Zhen, Peihao Chen, Shuhong Zheng, et~al.,
\newblock ``3d-llm: Injecting the 3d world into large language models,''
\newblock {\em Advances in Neural Information Processing Systems}, vol. 36, pp. 20482--20494, 2023.

\bibitem{3dmit}
Zeju Li, Chao Zhang, Xiaoyan Wang, Ruilong Ren, Yifan Xu, Ruifei Ma, Xiangde Liu, and Rong Wei,
\newblock ``3dmit: 3d multi-modal instruction tuning for scene understanding,''
\newblock in {\em 2024 IEEE International Conference on Multimedia and Expo Workshops (ICMEW)}. IEEE, 2024, pp. 1--5.

\bibitem{xu2025argus}
Yifan Xu, Chao Zhang, Hanqi Jiang, Xiaoyan Wang, Ruifei Ma, Yiwei Li, Zihao Wu, Zeju Li, and Xiangde Liu,
\newblock ``Argus: Leveraging multiview images for improved 3-d scene understanding with large language models,''
\newblock {\em arXiv preprint arXiv:2507.12916}, 2025.

\bibitem{huang2023chat3dv2}
Haifeng Huang, Zehan Wang, Rongjie Huang, Luping Liu, Xize Cheng, et~al.,
\newblock ``Chat-3d v2: Bridging 3d scene and large language models with object identifiers,''
\newblock {\em arXiv preprint arXiv:2312.08168}, 2023.

\bibitem{huang2024leo}
Jiangyong Huang, Silong Yong, Xiaojian Ma, Xiongkun Linghu, et~al.,
\newblock ``An embodied generalist agent in 3d world,''
\newblock in {\em Forty-first International Conference on Machine Learning}, 2024.

\bibitem{gao2024physically}
Jensen Gao, Bidipta Sarkar, Fei Xia, Ted Xiao, Jiajun Wu, Brian Ichter, Anirudha Majumdar, and Dorsa Sadigh,
\newblock ``Physically grounded vision-language models for robotic manipulation,''
\newblock in {\em 2024 IEEE International Conference on Robotics and Automation}. IEEE, 2024, pp. 12462--12469.

\bibitem{konenkov2024vr}
Mikhail Konenkov, Artem Lykov, Daria Trinitatova, and Dzmitry Tsetserukou,
\newblock ``Vr-gpt: Visual language model for intelligent virtual reality applications,''
\newblock {\em arXiv preprint arXiv:2405.11537}, 2024.

\bibitem{yang2024llplace}
Yixuan Yang, Junru Lu, Zixiang Zhao, et~al.,
\newblock ``Llplace: The 3d indoor scene layout generation and editing via large language model,''
\newblock {\em arXiv preprint arXiv:2406.03866}, 2024.

\bibitem{chen2024ll3da}
Sijin Chen, Xin Chen, Chi Zhang, Mingsheng Li, et~al.,
\newblock ``Ll3da: Visual interactive instruction tuning for omni-3d understanding reasoning and planning,''
\newblock in {\em Proceedings of the IEEE/CVF Conference on Computer Vision and Pattern Recognition}, 2024, pp. 26428--26438.

\bibitem{chen2024grounded3dllm}
Yilun Chen, Shuai Yang, Haifeng Huang, Tai Wang, et~al.,
\newblock ``Grounded 3d-llm with referent tokens,''
\newblock {\em arXiv preprint arXiv:2405.10370}, 2024.

\bibitem{cheng2024spatialrgpt}
An-Chieh Cheng, Hongxu Yin, Yang Fu, Qiushan Guo, Ruihan Yang, and ohters,
\newblock ``Spatialrgpt: Grounded spatial reasoning in vision language model,''
\newblock {\em arXiv preprint arXiv:2406.01584}, 2024.

\bibitem{chen2024spatialvlm}
Boyuan Chen, Zhuo Xu, Sean Kirmani, Brain Ichter, Dorsa Sadigh, et~al.,
\newblock ``Spatialvlm: Endowing vision-language models with spatial reasoning capabilities,''
\newblock in {\em Proceedings of the IEEE/CVF Conference on Computer Vision and Pattern Recognition}, 2024, pp. 14455--14465.

\bibitem{li2023m3dbench}
Mingsheng Li, Xin Chen, Chi Zhang, Sijin Chen, Hongyuan Zhu, Fukun Yin, Gang Yu, and Tao Chen,
\newblock ``M3dbench: Let's instruct large models with multi-modal 3d prompts,''
\newblock {\em arXiv preprint arXiv:2312.10763}, 2023.

\bibitem{lyu2024mmscan}
Ruiyuan Lyu, Tai Wang, Jingli Lin, Shuai Yang, Xiaohan Mao, Yilun Chen, Runsen Xu, Haifeng Huang, Chenming Zhu, Dahua Lin, et~al.,
\newblock ``Mmscan: A multi-modal 3d scene dataset with hierarchical grounded language annotations,''
\newblock {\em arXiv preprint arXiv:2406.09401}, 2024.

\bibitem{azuma2022scanqa}
Daichi Azuma, Taiki Miyanishi, Shuhei Kurita, and Motoaki Kawanabe,
\newblock ``Scanqa: 3d question answering for spatial scene understanding,''
\newblock in {\em proceedings of the IEEE/CVF conference on computer vision and pattern recognition}, 2022, pp. 19129--19139.

\bibitem{ma2022sqa3d}
Xiaojian Ma, Silong Yong, Zilong Zheng, Qing Li, Yitao Liang, Song-Chun Zhu, and Siyuan Huang,
\newblock ``Sqa3d: Situated question answering in 3d scenes,''
\newblock {\em arXiv preprint arXiv:2210.07474}, 2022.

\bibitem{zhang2023multi3drefer}
Yiming Zhang, ZeMing Gong, and Angel~X Chang,
\newblock ``Multi3drefer: Grounding text description to multiple 3d objects,''
\newblock in {\em Proceedings of the CVPR}, 2023, pp. 15225--15236.

\bibitem{chen2020scanrefer}
Dave~Zhenyu Chen, Angel~X Chang, and Matthias Nie{\ss}ner,
\newblock ``Scanrefer: 3d object localization in rgb-d scans using natural language,''
\newblock in {\em European conference on computer vision}. Springer, 2020, pp. 202--221.

\bibitem{chen2021scan2cap}
Zhenyu Chen, Ali Gholami, et~al.,
\newblock ``Scan2cap: Context-aware dense captioning in rgb-d scans,''
\newblock in {\em Proceedings of the IEEE/CVF conference on computer vision and pattern recognition}, 2021, pp. 3193--3203.

\bibitem{dai2017scannet}
Angela Dai, Angel~X Chang, Manolis Savva, Maciej Halber, Thomas Funkhouser, and Matthias Nie{\ss}ner,
\newblock ``Scannet: Richly-annotated 3d reconstructions of indoor scenes,''
\newblock in {\em Proceedings of the IEEE/CVF Conference on Computer Vision and Pattern Recognition}, 2017.

\bibitem{qi2017pointnet++}
Charles~Ruizhongtai Qi, Li~Yi, Hao Su, and J~Guibas,
\newblock ``Pointnet++: Deep hierarchical feature learning on point sets in a metric space,''
\newblock {\em Advances in neural information processing systems}, vol. 30, 2017.

\bibitem{votenet}
Charles~R Qi, Or~Litany, Kaiming He, and J~Guibas,
\newblock ``Deep hough voting for 3d object detection in point clouds,''
\newblock in {\em proceedings of the IEEE/CVF International Conference on Computer Vision}, 2019, pp. 9277--9286.

\bibitem{Vote2cap-detr++}
Sijin Chen, Hongyuan Zhu, Mingsheng Li, Xin Chen, Peng Guo, Yinjie Lei, YU~Gang, Taihao Li, and Tao Chen,
\newblock ``Vote2cap-detr++: Decoupling localization and describing for end-to-end 3d dense captioning,''
\newblock {\em IEEE Transactions on Pattern Analysis and Machine Intelligence}, 2024.

\bibitem{vedantam2015cider}
Ramakrishna Vedantam, C~Lawrence~Zitnick, and Devi Parikh,
\newblock ``Cider: Consensus-based image description evaluation,''
\newblock in {\em Proceedings of CVPR}, 2015, pp. 4566--4575.

\bibitem{papineni2002bleu}
Kishore Papineni, Salim Roukos, Todd Ward, and Wei-Jing Zhu,
\newblock ``Bleu: a method for automatic evaluation of machine translation,''
\newblock in {\em Proceedings of the 40th annual meeting of the Association for Computational Linguistics}, 2002, pp. 311--318.

\bibitem{banerjee2005meteor}
Satanjeev Banerjee and Alon Lavie,
\newblock ``Meteor: An automatic metric for mt evaluation with improved correlation with human judgments,''
\newblock in {\em Proceedings of the acl workshop on intrinsic and extrinsic evaluation measures for machine translation and/or summarization}, 2005, pp. 65--72.

\bibitem{lin2004rouge}
Chin-Yew Lin,
\newblock ``Rouge: A package for automatic evaluation of summaries,''
\newblock in {\em Text summarization branches out}, 2004, pp. 74--81.

\bibitem{3dvista}
Ziyu Zhu, Xiaojian Ma, Yixin Chen, Zhidong Deng, Siyuan Huang, and Qing Li,
\newblock ``3d-vista: Pre-trained transformer for 3d vision and text alignment,''
\newblock in {\em Proceedings of the IEEE/CVF International Conference on Computer Vision}, 2023, pp. 2911--2921.

\bibitem{Yang2023LLMGrounderO3}
Jianing Yang, Xuweiyi Chen, Shengyi Qian, Nikhil Madaan, et~al.,
\newblock ``Llm-grounder: Open-vocabulary 3d visual grounding with large language model as an agent,''
\newblock {\em 2024 IEEE International Conference on Robotics and Automation}, pp. 7694--7701, 2023.

\bibitem{Zhu2024ScanReasonE3}
Chenming Zhu, Tai Wang, Wenwei Zhang, Kai Chen, and Xihui Liu,
\newblock ``Scanreason: Empowering 3d visual grounding with reasoning capabilities,''
\newblock {\em ArXiv}, vol. abs/2407.01525, 2024.

\bibitem{absolute_relative_error}
Kani Chen, Shaojun Guo, Yuanyuan Lin, and Zhiliang Ying,
\newblock ``Least absolute relative error estimation,''
\newblock {\em Journal of the American Statistical Association}, vol. 105, no. 491, pp. 1104--1112, 2010.

\bibitem{zhang2022opt}
Susan Zhang, Stephen Roller, Naman Goyal, Mikel Artetxe, Moya Chen, Shuohui Chen, Christopher Dewan, Mona Diab, Xian Li, Xi~Victoria Lin, et~al.,
\newblock ``Opt: Open pre-trained transformer language models,''
\newblock {\em arXiv preprint arXiv:2205.01068}, 2022.

\bibitem{touvron2023llama}
Hugo Touvron, Louis Martin, Kevin Stone, Peter Albert, Amjad Almahairi, Yasmine Babaei, Nikolay Bashlykov, Soumya Batra, Prajjwal Bhargava, Shruti Bhosale, et~al.,
\newblock ``Llama 2: Open foundation and fine-tuned chat models,''
\newblock {\em arXiv preprint arXiv:2307.09288}, 2023.

\end{thebibliography}

\appendix

\section{supplementary material}
In this supplementary material, we first provide examples related to the proposed dataset MODEL, and outline the evaluation metrics. Following this, we report more evaluations and the qualitative experimental results. These details are discussed in the subsequent sections.

\subsection{Datasets}


\subsubsection{Example of 3D object distance measurement task}
\begin{itemize}
    \item \textbf{Scene:} \texttt{scene0011\_00}
    \item \textbf{Question:} \textit{"Object A is described as: 'There is a set of bottom kitchen cabinets in the room. It has a microwave in the middle of it.' Object B is described as: 'You are looking for a chair on the side of the table facing the ovens. It will be the chair near the rail.' Please provide the distance between Object A and Object B."}
    \item \textbf{Answer:} \textit{"Object A is a kitchen\_cabinets located at \texttt{<loc>198, 171, 47, 7, 96, 81</loc>}. Object B is a chair located at \texttt{<loc>141, 110, 58, 21, 16, 96</loc>}. The spatial distance from Object A to Object B on the x-axis is \texttt{<gap>57</gap>} units, on the y-axis is \texttt{<gap>61</gap>} units, and on the z-axis is \texttt{<gap>11</gap>} units."}
\end{itemize}
\subsubsection{Evaluation metrics of 3D object distance measurement task}
To assess the accuracy of distance predictions, we draw inspiration from the evaluation of 3D-VG, focusing on the localization accuracy of Objects A and B. We follow the approach used in \cite{absolute_relative_error}, employing \textbf{absolute relative error (ARE)} to evaluate distance prediction accuracy. Our primary metric, \textbf{mARE@kIOU}, measures the mean absolute relative error for predictions on the X, Y, and Z axes, providing a detailed assessment of the model's spatial reasoning capabilities.

\subsubsection{Example of 3D layout editing task}
\begin{itemize}
    \item \textbf{Object Movement - Scene:} \texttt{scene0011\_00}
    \item \textbf{Question:} \textit{"Based on the provided description, 'this is a brown cabinet, it sets along the wall, right next to a window.' Move the object that closely matches this description forward by 110 units, and then describe its new location."}
    \item \textbf{Answer:} \textit{"It is a cabinet located at \texttt{<loc>209, 61, 160, 27, 32, 153</loc>}. Its location after moving forward by 110 units is \texttt{<loc>209, 171, 160, 27, 32, 153</loc>}."}
\end{itemize}

\begin{itemize}
    \item \textbf{Object Placement - Scene:} \texttt{scene0011\_00}
    \item \textbf{Question:} \textit{"Add a chair with size w:27, h:22, l:96 to the current indoor scene, and please output the center coordinates of the object."}
    \item \textbf{Answer:} \textit{"133, 80, 57"}
\end{itemize}
\subsubsection{Evaluation metrics of 3D layout editing task}
To evaluate the accuracy of object editing in the scene, we follow the metrics used in the 3D-VG task, calculating the Intersection over Union (IoU) between the predicted bounding box and the ground truth to assess the rationality of the predicted positions.

\subsection{More Evaluations}

To analyze the impact of individual tasks on the overall performance of the model, Table\ref{tab: ablation-tasks} reports performance under different training setups:3D VL understanding (\textbf{U}), 3D VL grounding (\textbf{G}), and our proposed task (\textbf{O}). These findings underscore the importance of joint training in improving 3D scene understanding and spatial perception.

To further evaluate the effectiveness of our model, we conducted training and evaluation on embodiedQA tasks. Specifically, we employed the Scene Description, Embodied Dialogue, and Embodied Planning tasks from the LL3DA\cite{chen2024ll3da}. As shown in Table\ref{tab: EQA-tasks} under quantitative comparisons, our method substantially outperformed existing approaches in both CiDEr and METEOR scores, demonstrating superior capability in generating diverse and contextually appropriate language.

\begin{table}
\centering
\caption{\textbf{Ablation studies of different tasks.} \textbf{U} refer to train on 3D VL understanding tasks, \textbf{G} refer to training on 3D VL grounded tasks, and \textbf{O} refer to training on our proposed task.}

\label{tab: ablation-tasks}
\resizebox{\linewidth}{!}{
\begin{tabular}{lccccccccccccccccccccc} 
\toprule
& \multicolumn{2}{c}{Scan2Cap} &  & \multicolumn{2}{c}{Multi3DRef Grd} &  & Movement &  & Placement &  & Measurement       \\ 
\cmidrule{2-3}\cmidrule{5-6}\cmidrule{8-8}\cmidrule{10-10}\cmidrule{12-12}
                  & C        & B-4      &  & F1@0.25 & F1@0.5      &  & Acc@0.5   &  & Acc@0.5  &  & X/Y/Z-mARE@0.5  \\ 
\midrule
Ours (U only)     & 66.9    & 33.0      &  & -       & -           &  & -         &  & -        &  & -           \\
Ours (G only)     &    -    & -         &  & 45.3    & 40.3        &  & -         &  & -        &  & -        \\
Ours (G + O)      & -       & -         &  & 47.2    & 40.1        &  & 37.5      &  & 64.4     &  & 2.5/2.1/3.1         \\ 
Ours (U + G)      & 68.2    & 34.2      &  & 48.3    & 42.7        &  & -         &  & -        &  & -           \\ 
Ours (U + G + O)  & 72.9    & 35.5      &  & 49.0    & 43.5        &  & 41.1      &  & 70.0     &  & 2.0/1.8/2.1        \\
\bottomrule
\end{tabular}}
\end{table}

\begin{table*}[!tb]
\centering
\caption{Quantitative comparison with SOTA models on embodiedQA tasks.}
\label{tab: EQA-tasks}
\resizebox{\linewidth}{!}{
\begin{tabular}{lllllllll} 
\toprule
Task              &  Method         & BLEU-1$\uparrow$ & BLEU-2$\uparrow$ & BLEU-3$\uparrow$ & BLEU-4$\uparrow$ & CiDEr$\uparrow$  & METEOR$\uparrow$ & Rouge-L$\uparrow$  \\ 
\midrule
Scene Description & \textbf{Zero-Shot:} &        &        &        &        &        &        &          \\
                  & OPT-1.3B\cite{zhang2022opt}        & 15.79  & 6.10   & 2.07   & 0.84   & 0.00   & 8.40   & 11.70    \\
                  & OPT-2.7B\cite{zhang2022opt}        & 19.97  & 7.59   & 2.14   & 0.00   & 0.11   & 6.60   & 12.32    \\
                  & OPT-6.7B\cite{zhang2022opt}        & 24.40  & 9.79   & 3.62   & 1.13   & 0.06   & 8.99   & 16.96    \\
                  & LLAMA-7B\cite{touvron2023llama}        & 19.26  & 7.69   & 2.79   & 0.92   & 0.20   & 7.00   & 12.31    \\ 
\cmidrule{2-9}
                  & \textbf{Ours}       &        &        &        &        &        &        &          \\
                  & LL3DA\cite{chen2024ll3da}    & 43.02  & 26.7   & 15.97  & 8.97   & 0.96   & 14.65  & 24.84    \\
                  & Ours                & 42.67  & 25.03  & 15.29  & 9.37   & 5.50   & 17.52  & 23.88    \\ 
\midrule
Embodied Dialogue & \textbf{Zero-Shot:} &        &        &        &        &        &        &          \\
                  & OPT-1.3B\cite{zhang2022opt}        & 2.44   & 1.05   & 0.46   & 0.23   & 0.31   & 5.62   & 4.83     \\
                  & OPT-2.7B\cite{zhang2022opt}        & 3.88   & 1.56   & 0.73   & 0.39   & 0.38   & 7.38   & 6.28     \\
                  & OPT-6.7B\cite{zhang2022opt}        & 3.59   & 1.65   & 0.81   & 0.43   & 0.25   & 6.88   & 6.16     \\
                  & LLAMA-7B\cite{touvron2023llama}        & 4.08   & 1.80   & 0.90   & 0.50   & 0.27   & 7.81   & 6.68     \\ 
\cmidrule{2-9}
                  & \textbf{Ours}       &        &        &        &        &        &        &          \\
                  & LL3DA\cite{chen2024ll3da}     & 41.38  & 32.59  & 27.47  & 23.95  & 190.01 & 23.50  & 40.61    \\
                  & Ours                & 52.80  & 43.42  & 37.68  & 33.56  & 290.37 & 27.46  & 51.54    \\ 
\midrule
Embodied Planning & \textbf{Zero-Shot:} &        &        &        &        &        &        &          \\
                  & OPT-1.3B\cite{zhang2022opt}        & 1.26   & 0.59   & 0.26   & 0.13   & 0.16   & 0.24   & 3.56     \\
                  & OPT-2.7B\cite{zhang2022opt}        & 2.02   & 0.99   & 0.49   & 0.26   & 0.10   & 3.59   & 4.35     \\
                  & OPT-6.7B\cite{zhang2022opt}        & 2.03   & 1.06   & 0.53   & 0.28   & 0,00   & 3.65   & 3.94     \\
                  & LLAMA-7B\cite{touvron2023llama}        & 2.24   & 1.13   & 0.55   & 0.29   & 0.04   & 3.53   & 4.71     \\ 
\cmidrule{2-9}
                  & \textbf{Ours}       &        &        &        &        &        &        &          \\
                  & LL3DA\cite{chen2024ll3da}    & 40.72  & 27.18  & 18.64  & 12.95  & 128.80 & 17.05  & 39.25    \\
                  & Ours                & 37.35  & 26.42  & 19.56  & 14.65  & 195.92 & 18.95  & 36.93    \\
\bottomrule
\end{tabular}}
\end{table*}

\section{qualitative experimental results of our method}
\begin{figure*}[t]
  \centering
  \includegraphics[width=0.9\textwidth]{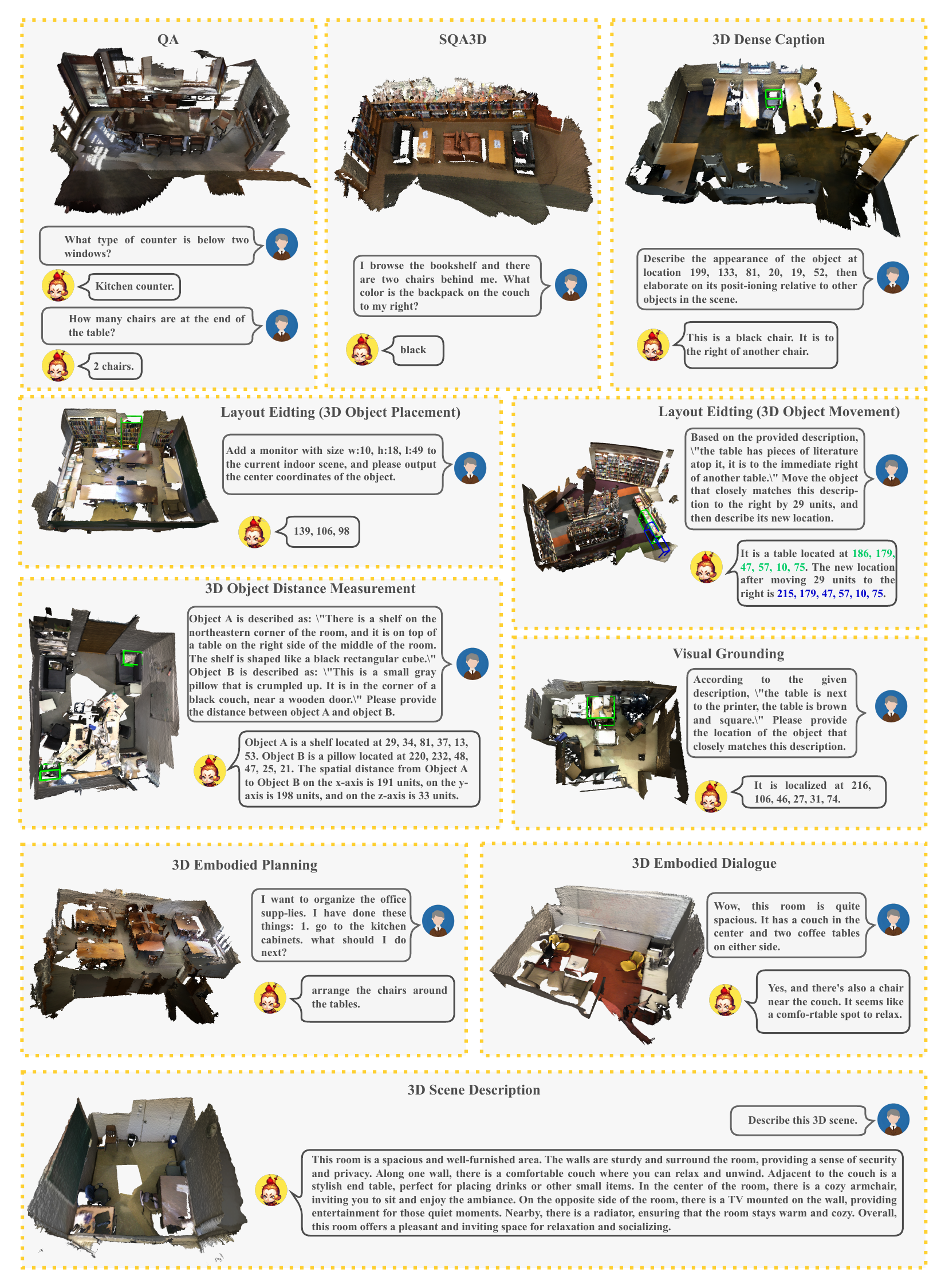}
  \caption{\textbf{Qualitative Results.} We provide several visualization results on various 3D vision and language tasks.}
  \label{fig:case1}
\end{figure*}

Figure~\ref{fig:case1} shows the qualitative results used to evaluate the spatial awareness capabilities of our model.

\end{document}